## 1. Overview

**Title**
COVID-19 case data for Italy stratified by age class

**Paper Authors**
1. Giuseppe Calafiore; (Lead author and corresponding author)
2. Giulia Fracastoro

**Authors' Roles and Affiliations**
1. Professor, Dipartimento di Elettronica e Telecomunicazioni, Politecnico di Torino, Italy;
2. Researcher, Dipartimento di Elettronica e Telecomunicazioni, Politecnico di Torino, Italy.
Email: giuseppe.calafiore@polito.it; giulia.fracastoro@polito.it

**Abstract**
The dataset described in this paper contains daily data about COVID-19 cases that occurred in Italy over the period from Jan. 28, 2020 to March 20, 2021, divided into ten age classes of the population, the first class being 0-9 years, the tenth class being >90 years.  The dataset contains eight columns, namely: date (day), age class, number of new cases, number of newly hospitalized patients, number of patients entering intensive care, number of deceased patients, number of recovered patients, number of active infected patients. This data has been officially released for research purposes by the Italian authority for COVID-19 epidemiologic surveillance (*Istituto Superiore di Sanità* – ISS), upon formal request by the authors, in accordance with the Ordonnance of the Chief of the Civil Protection Department n. 691 dated Aug. 4 2020. A separate file contains the numerosity of the population in each age class, according to the *National Institute of Statistics'* (ISTAT) data of the resident population of Italy as of Jan. 2020. This data has potential use, for instance, in epidemiologic studies of the effects of the COVID-19 contagion in Italy, in mortality analysis by age class, and in the development and testing of dynamical models of the contagion.

**Introduction**
The age structure of the population appears to play a key role in determining the severity of symptoms and the mortality of the disease caused by the SARS-CoV-2 infection. The importance of the demographic structure in determining the pandemic's progression and impact has indeed been well recognized by researchers, see, e.g., [1], [2]. Also, a clearer understanding of the contagion's interaction dynamics among age classes appears to be fundamental for devising effective containment measures and for establishing priorities for the vaccination campaigns. Despite the importance of age-related COVID-19 data, and despite the fact that calls for countries to provide this data have been repeatedly made (see, e.g., [2], [3], [4]) this type of data has been to date essentially unavailable to the public, and even to researchers. This fact motivated us to formally request specific age-related COVID-19 data to Italian authorities in charge of the COVID-19 surveillance (*Istituto Superiore di Sanità* – ISS), so to make them available to the public for research purposes.

## 2. Context and Methods

**Spatial and temporal coverage**
The data refers to the population of Italy and covers the period from Jan. 28, 2020 to March 20, 2021, with daily frequency. Data relative to the early phase of the contagion (i.e., previous to March 2020) have several missing values for some age classes.

**Methodology and quality control**
The data reported in the file are the data present in the Italian COVID-19 surveillance system, updated to the extraction date of March 22, 2021. The data represents aggregations of positive cases for SARS-CoV-2 derived from the Integrated Covid-19 Surveillance coordinated by the ISS (Ordonnance no. 640 of February 27, 2020). The Integrated Surveillance data is updated daily by each Region, both with new cases and with the addition of new information on cases already communicated previously, as they become available. In addition, the constant quality control of the data also seldom highlights the need, on the part of the Regions, to cancel some cases that are mistakenly duplicated.
The data collected is in a continuous phase of consolidation and, as expected in an emergency situation, some information is incomplete. In particular, the possibility of a delay of a few days between the execution of the swab for diagnosis and reporting on the dedicated platform is noted. Therefore, the number of cases observed in the most recent days, compared to the extraction date, must be interpreted as provisional and incomplete. The same applies to reporting hospitalization and death.

**Privacy**
The data reported are disaggregated in a manner that guarantees compliance with the privacy legislation. In particular, it should be noted that for frequency values between 1 and 4 the value is expressed as "<5".

## 4. Dataset description

**Object name**
Two files are provided. The first file is the main COVID-19 data file named "covid_ageclass_Italy.csv" while the second file named "ageclass_pop.csv" is an ancillary file that contains the population cardinality for each age class.

**Format names and versions**
File format is textual comma separated values (CSV).

**Dataset creators**
The dataset was extracted from the national official database on March 22, 2021, upon request from the authors, by Dr. Patrizio Pezzotti from the Epidemiology, Biostatistics and Mathematical Models Department of the ISS.

**Licence**
The data is provided under the CC0-Public Domain Dedication waiver licence.

**Repository location**
dataverse.harvard.edu: https://doi.org/10.7910/DVN/VSS4CO

**Publication date:** April 13, 2021.

**Data description**
The data file "covid_ageclass_Italy.csv" contains 4015 rows (plus the headings row) and eight columns. The columns contain the following data:

1. "date" contains the date indicating the day to which the data in the other columns refers. It is the date of the confirmed diagnosis of microbiological SARS-CoV-2 infection, or the date of hospitalization, the date of recovery, the date of death, etc.
2. "age_class" is the age class, in a ten-year range. In some rare cases it can be "Unknown."
3. "cases" contains the number of confirmed positive SARS-CoV-2 infected cases for that day in the given age class.
4. "hospitalized" contains the number of patients hospitalized (due to COVID) in that day in the given age class.
5. "intensive_care" contains the number of patients that entered intensive care (due to COVID) in that day in the given age class.
6. "deceased" contains the number of deceased persons (with death ascribed to COVID) in that day in the given age class.
7. "recovered" contains the number of persons that recovered (from COVID) in that day in the given age class.
8. "active_infected" contains the total number of persons that are active and infected with SARS-CoV-2 on the given day in the given age class.

The ancillary file data file "ageclass_pop.csv" contains 10 rows (plus the headings row) and two columns. The first column "age_class" contains the age class, the second column "population" contains the number of individuals resident in Italy for that age class, as of Jan. 2020.

**Data overview**
A cumulative summary of part of the data is shown in Table 1. Mortality is here computed simply as the ratio between deceased individuals in a given age class and the population of that class. Lethality is computed as the ratio between deceased individuals in a given age class and the infected individuals (cases) in that class. Values reported as "<5" in the data are imputed a default value of 2. Figure 1 shows a pie chart of the deaths by age. Figure 2 shows an example of time-series data representing the daily cases for the 50-59 age class; the regular spikes in the plot correspond to Sundays. Figure 3 shows the time-series of the active infected individuals for the 50-59 age class; three infection peaks are visible, the first in mid-April 2020, the second in late November 2020, and the third in formation mid-March 2021.

Table 1: summary table.

| Age class | Population | Cases | Intensive care | Deceased | %Mortality | %Lethality |
|-----------|-----------|-------|----------------|----------|-----------|-----------|
| **0-9** | 4892494 | 155418 | 154 | 22 | 0.0005 | 0.0142 |
| **10-19** | 5706116 | 295710 | 169 | 24 | 0.0004 | 0.0081 |
| **20-29** | 6084382 | 386598 | 378 | 90 | 0.0015 | 0.0233 |
| **30-39** | 6854632 | 403558 | 834 | 248 | 0.0036 | 0.0615 |
| **40-49** | 8937229 | 521716 | 2451 | 829 | 0.0093 | 0.1589 |
| **50-59** | 9414195 | 573800 | 6733 | 3362 | 0.0357 | 0.5859 |
| **60-69** | 7364364 | 362317 | 11425 | 9700 | 0.1317 | 2.6772 |
| **70-79** | 5968373 | 268141 | 13437 | 24866 | 0.4166 | 9.2735 |
| **80-89** | 3628160 | 216635 | 7467 | 42546 | 1.1727 | 19.6395 |
| **>=90** | 791543 | 77721 | 1360 | 20869 | 2.6365 | 26.8512 |
| **ALL** | **59641488** | **3261614** | **44408** | **102556** | **0.1720** | **3.1443** |

Figure 1: deaths by age class.

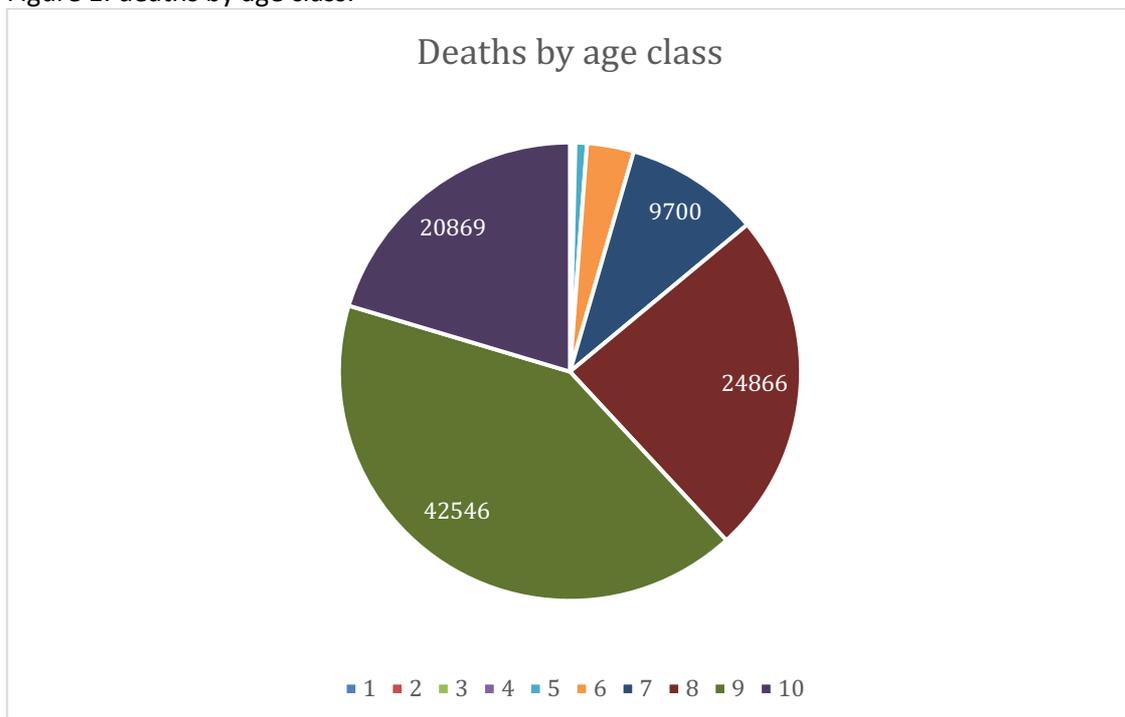

Figure 2: cases for the 50-59 age class.

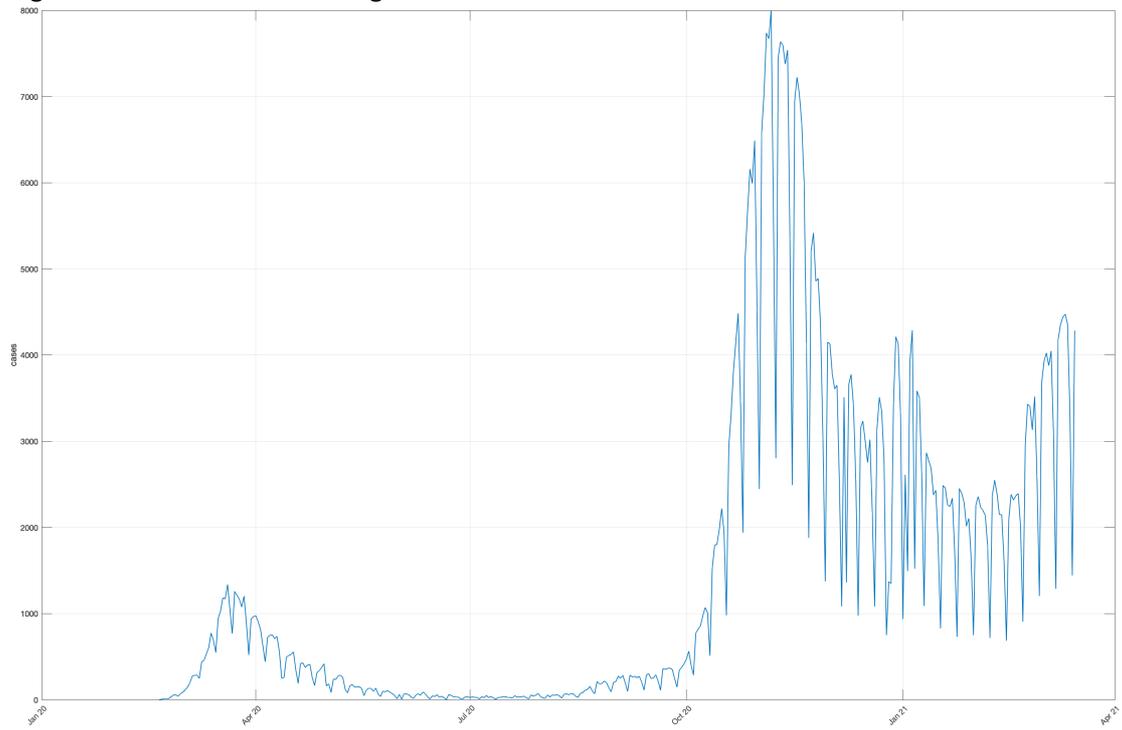

Figure 3: active infected individuals for the 50-59 age class.

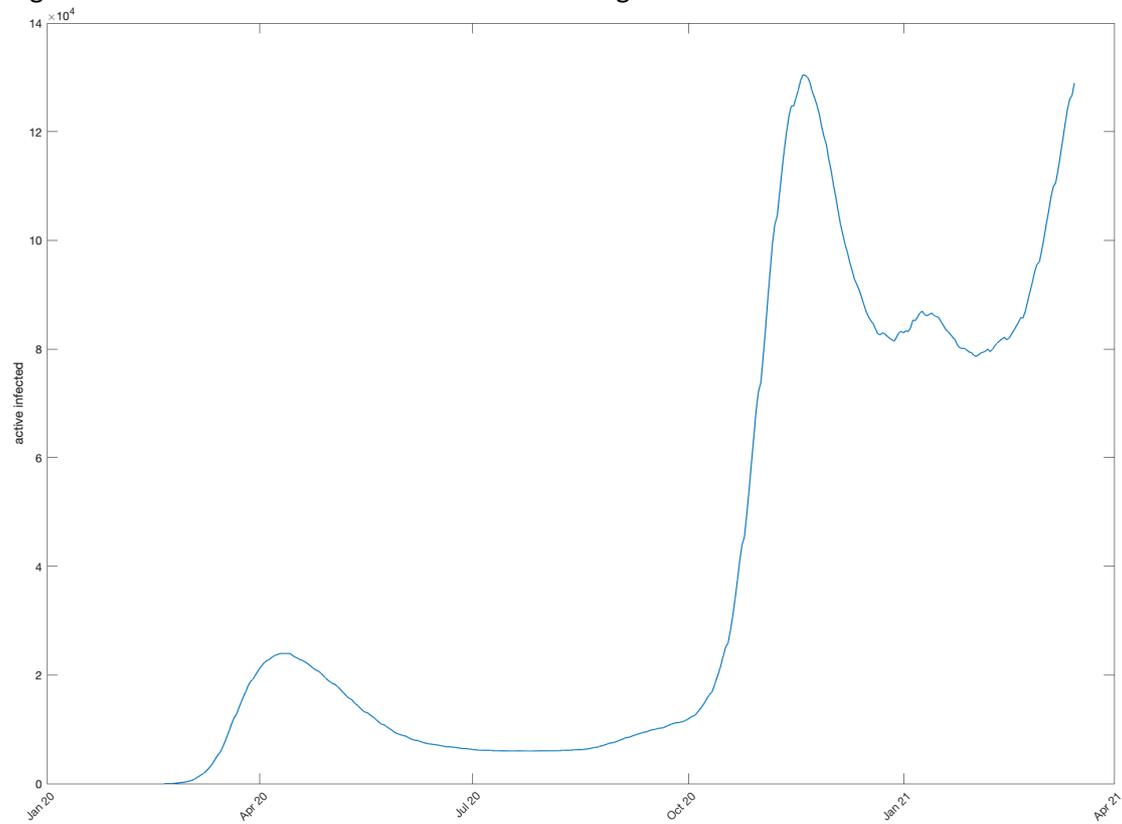

**5. Reuse potential**

The data can be used for research purposes, including aggregation, analysis, reference, model (e.g., SIRD) building and validation, teaching or collaboration.

**Acknowledgements**

We acknowledge the help of prof. Andrea Bianco, head of the Department of Electronics and Telecommunications Engineering of Politecnico di Torino, Italy, for his help in managing the formal data request to ISS.